\documentclass[10pt,twocolumn,letterpaper]{article}

\usepackage{iccv}
\usepackage{times}
\usepackage{epsfig}
\usepackage{graphicx}
\usepackage{import} 
\usepackage{amsmath}
\usepackage{amssymb}
\usepackage{verbatim}
\usepackage{tabularx}
\usepackage{multirow}
\usepackage{appendix}
\usepackage{csquotes}

\usepackage[pagebackref=true,breaklinks=true,letterpaper=true,colorlinks,bookmarks=false]{hyperref}

\iccvfinalcopy 

\def\iccvPaperID{****} 

\ificcvfinal\fi

\begin{document}

\title{PatFig: Generating Short and Long Captions for Patent Figures}

\author{
    Dana Aubakirova\textsuperscript{1,2} \and 
    Kim Gerdes\textsuperscript{1,2} \and 
    Lufei Liu\textsuperscript{1} \and
    \\
    \textsuperscript{1}Qatent, Paris, France 
    \textsuperscript{2}Université Paris-Saclay, LISN (CNRS), France \\
    \{dana, kim, lufei\}@qatent.com
}



\maketitle
\ificcvfinal\thispagestyle{empty}\fi

\begin{abstract}

This paper introduces Qatent PatFig, a novel large-scale patent figure dataset comprising 30,000+ patent figures from over 11,000 European patent applications.
For each figure, this dataset
provides short and long captions, reference numerals, their corresponding terms, and the minimal claim set that describes the interactions between the components of the image.
To assess the usability of the dataset, we finetune an LVLM model on Qatent PatFig to generate short and long descriptions,
and we investigate the effects of incorporating various text-based cues at the prediction stage of the patent figure captioning process. 
\end{abstract}

\section{Introduction}
Patents are at the economically strategic crossroads of Artificial Intelligence and Intellectual Property, serving as a cornerstone of technical innovation\cite{campbell1983patent}. A pivotal yet largely untapped aspect at the confluence of visual and linguistic analysis is the study of patent figures. These figures are central to the comprehension and elucidation of patent applications, often providing a more efficient medium for conveying complex scientific or technical information than text alone \cite{carney2002pictorial, ganguly2011united}. They comprise technical drawings, block diagrams, flow charts, plots, and grayscale photographs \cite{wei2022visual}. 
While prior research has delved into captioning scientific figures, the specific domain of patent figure captioning remains largely unexplored.
\begin{figure}[t]
\begin{center}
   \includegraphics[width=0.8\linewidth]{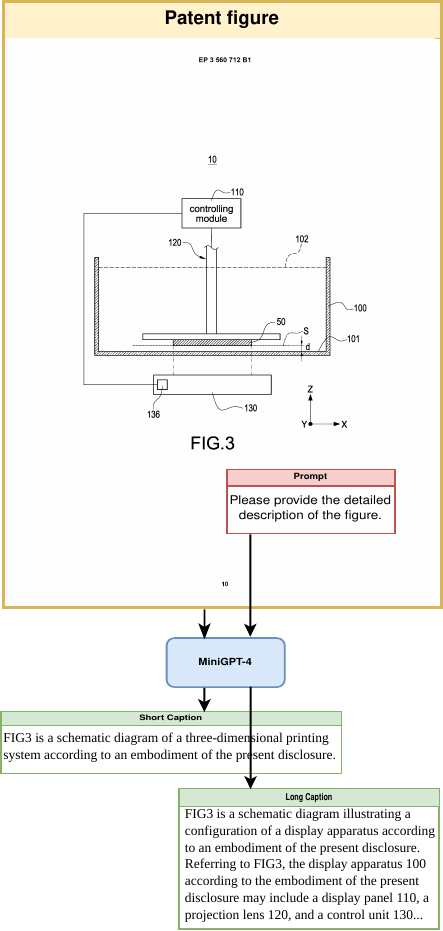}
  \caption{Given an image and a prompt, our figure captioning models generate long and short descriptions. Note that the models are separate for the two types of descriptions.
  \vspace{-12mm}}
  \end{center}
\label{fig:onecol}
\end{figure}
We introduce Qatent PatFig, a comprehensive patent figure dataset with long and short descriptions, bolstering research in areas like image-to-text, figure-based patent retrieval, figure classification, segmentation, and text-to-image generation. 
Using PatFig, we train image captioning models to aid patent attorneys in improving figure captions. 
By fine-tuning the Large Vision Language Model MiniGPT-4 \cite{zhu2023minigpt} on PatFig and adding textual cues during predictions, we strive to boost caption accuracy and patent-specific relevance.
The ultimate goal is to connect vision and language, by comprehending visual information designed to facilitate human cognition of abstract, technical, and scientific concepts. This endeavor is particularly relevant in the context of patent figures, which often encapsulate complex and abstract concepts in a visual format \cite{jones2022abstract}. 

\section{Related work}
\subsection{Patent figure datasets}
Limited datasets exist for patent figure analysis, primarily targeting image-based patent retrieval. CLEF-IP 2011 \cite{piroi2011clef} provides two such datasets, but with a mere 211 patents and broad image classification across nine categories, it is limited in granularity. The concept dataset \cite{zhang2015patent} has 1000 patent drawings for shoe classification and an additional 2000 mechanical drawings by relevance. Kucer et al.'s DeepPatent \cite{kucer2022deeppatent} offers over 350,000 design patent images\footnote{In Europe, "design patents" are termed "Registered Community Design" (RCD). They focus on aesthetic design rather than utility.}. Such patents naturally lack detailed object names, viewpoints, and captions.

We introduce Qatent PatFig, a comprehensive dataset with 30,000 patent figures from 11,000+ patents enriched with long and short descriptions, figure types, reference numerals with terms, and patent claims. While similar datasets like SciCap \cite{hsu2021scicap} associate scientific figures with captions, patent figures present unique challenges. They frequently feature reference numerals, term lists, short as well as long descriptions, and the relation between the terms is detailed in the patent claims. However, extracting these descriptions can be daunting due to varied caption structures and the interspersed nature of reference numerals throughout the patent.

\subsection{Patent figure captioning}
Most patent figure research targets figure-based patent querying \cite{kucer2022deeppatent, vrochidis2012concept, piroi2011clef} and classification \cite{jones2022abstract, zhang2022reliable, li2018deeppatent}. For scientific figure captioning, Chen et al. \cite{chen2019figure, chen2020figure, chen2019neural} presented FigCAP, using LSTM models with attention. Qian et al.'s FigJAM \cite{qian2021generating} produces \enquote{caption units}, a concept explored earlier with DVQA \cite{kafle2018dvqa} and FigureQA \cite{kahou2017figureqa}. SciCap \cite{hsu2021scicap} leverages its dataset to train a CNN+LSTM image-captioning model \cite{xu2015show}.


In this paper, we leverage the recent advancements in Large Vision-Language Models (LVLMs) to address the task of generating short and long captions for patent figures. LLMs, such as LLaMA \cite{touvron2023llama}, GPT-3 \cite{madotto2020language}, and Vicuna \cite{chiang2023vicuna} have demonstrated disruptive progress that can be further extended to large vision-language models \cite{alayrac2022flamingo, zhu2023minigpt, li2023blip}, thus effectively aligning visual features with the textual space. Yet, their application to the domain of patent figure captioning remains unexplored. We propose to finetune an LVLM to evaluate our dataset and investigate the effectiveness of LVLMs in generating informative and detailed captions for patent figures. 
\section{Building the PatFig dataset}
In this section, we describe the process of acquiring and pre-processing the data to construct our dataset. 

\subsection{Data acquisition and pre-processing}
Qatent's internal Solr \cite{solr} database contains complete textual patent data from the European Patent Office (EPO)
including publication number, title, abstract, claim, IPC (patent classification), inventors, patent family, applicants, id, and complete description. Based on this database, we initiated the image 
acquisition process by retrieving the publication numbers 
within the time range from January 1, 2020, to December 31, 2020. 
Subsequently, using Espacenet \cite{espacenet}, the EPO's patent search website, we scraped a total of 62,513 patent images corresponding to 15,645 unique patents based on the patent publication numbers, enabling for accurate linking of the images to their respective textual patent data. 

\subsection{Short and long figure caption extraction}

Short descriptions of patent figures usually follow a standard format, separated by new lines, enabling a rule-based extraction method. These descriptions often appear in a section titled \enquote{Brief Description of Drawings}. Typical short descriptions start with a figure number and a brief explanation, e.g., \textit{\enquote{Fig. 1 depicts a bottle pourer per an embodiment.}} This uniformity aids in automated extraction of such captions. Our results yielded structured sentences with figure numbers, objects, and viewpoints when available.

Long caption extraction poses more challenges due to the varied structure of patent application descriptions. Addressing this, our method involved text normalization, searching for repeated figure number references, and extracting relevant sections until the start of another paragraph or a different figure mention. We also trimmed overly verbose captions.\footnote{During caption filtering, statistical analysis determined token count ranges for descriptions. For short captions, the range was 10 to 40 tokens, based on the Interquartile Range (IQR) rule. For long ones, it was 40 to 500 tokens. Descriptions outside these bounds were treated as outliers and excluded.}

\subsection{Figure-type extraction}
We leverage the common structure of short captions and apply rule-based methods to extract key phrases appearing after the \enquote{are/is a/an,} \enquote{shows,} \enquote{illustrates,} and \enquote{depicts.}, etc. As a result, we retrieved 1506 different classes, which were later reduced to 412 after manual revision and text normalization. We grouped the most frequently appearing categories, ranging from the least abstract (top) to the most abstract (bottom) categories present in the dataset as illustrated in Figure \ref{fig:long} in Annex \ref{appendix:fig}. 

\subsection{Figure-caption matching with OCR}
OCR (Optical Character Recognition) is employed to extract reference numerals present in patent images, enabling their matching with corresponding figure descriptions and terms. We evaluated four OCR libraries: PyOCR, EasyOCR, Pytesseract, and docTR \cite{liao2023doctr} to determine the best method for extracting information from patent images. 
54 images from eight IPC patent class sections (A, B, C, D, E, F, G, H) were randomly selected and manually annotated with reference numerals and figure numbers. As many patent figures appear sideways, we tested rotating the images at 0, 90, 180, and 270 degrees. Among the libraries, we selected docTR for extracting the numerical information present in the image and matching it with the corresponding description as it achieved the highest accuracy of 72.08 as provided in Table \ref{tab:my_table1}. 

\begin{table}[]
\resizebox{\columnwidth}{!}{%
\begin{tabular}{l|llll}
\hline
 & PyOCR & EasyOCR & Pytesseract & docTR \\ \hline
Accuracy & 14.03 & 53.22 & 17.42 & 72.08 \\ 
Time (s) & 0.49 & 17.11 & 0.75 & 7.02 \\ \hline
\end{tabular}%
} \\
\caption{The OCR Performance}
\label{tab:my_table1}
\end{table}

\section{Dataset evaluation with LVLM}
While traditional approaches such as CNN+RNN models have been prevalent in image captioning tasks, they face limitations in effectively representing features and inhibit training parallelization due to the recurrent nature of RNNs \cite{wen2023vision}. Thus, given the domain-specific and technical nature of patent figures, relying solely on CNN+RNNs may not capture the intricate semantic relationships and contextual nuances required for generating accurate captions. Following the recent trends in 
utilizing autoregressive language models for vision-language tasks, capitalizing on cross-modal transfer, we chose a large vision-language image-captioning model, MiniGPT-4, as our baseline model. We created three variations of the baseline models, namely Vision-only and two versions of Vision+Text. We evaluate the generated short and long captions using common evaluation metrics: BLEU \cite{papineni2002bleu}, ROUGE\cite{lin2004rouge}, METEOR \cite{banerjee2005meteor}, and CIDEr \cite{vedantam2015cider} scores. 

\subsection{Experimental setup}
We finetuned the MiniGPT-4 model separately on short and long captions. For both models, we trained for 10 epochs, setting the maximum text length to 50 for short captions and 500 for long captions. The fine-tuning for short captions was carried out on a 1xRTX A6000 GPU, while the long captions were fine-tuned on a 1xA100 80GB GPU. MiniGPT-4 is composed of a vision encoder with a pre-trained Vision Transformer (ViT) and Q-Former, a single linear projection layer, and an advanced Vicuna large language model. Fine-tuning involved training only the linear projection layer to align the visual features with the Vicuna model. 

The authors of MiniGPT provide 2 stages of pretraining. In the first pretraining stage, MiniGPT-4 acquires vision\-language knowledge from a combined dataset of Conceptual Caption \cite{sharma2018conceptual, changpinyo2021conceptual}, SBU \cite{ordonez2011im2text}, and LAION \cite{schuhmann2021laion}. However, it exhibited issues with the coherence of the generated texts. To address this, they provide the second-stage finetuning on a smaller highly curated dataset to refine the generated descriptions. We fine\-tune on the second stage. The qualitative results are provided in \ref{tab:my_table4} in \ref{appendix:pred}. The fine-tuning process utilized predefined prompts such as: \enquote{Please provide the detailed description of the figure} or \enquote{Describe the contents of the image in detail.} The fine\-tuned MiniGPT-4 model on aligned patent figures with short and long captions exhibited the ability to produce more specific and contextually appropriate descriptions.\footnote{Notably, we observed that the fine\-tuning process was not efficient, requiring 4500 training steps with a batch size of 4, which is the maximum batch size suitable for the dataset with long captions (max. 500 tokens) that fits within a single 80GB A100 GPU. The multi\-GPU options for the fine\-tuning stage have not been released yet.}

\begin{table}[]
\resizebox{\columnwidth}{!}{%
\begin{tabular}{l|lll}
\hline
\multirow{4}{*}{\begin{tabular}[c]{@{}l@{}}The number of \\ images with captions\end{tabular}} & \multicolumn{1}{l|}{\multirow{2}{*}{\begin{tabular}[c]{@{}l@{}}Raw dataset \\ ($ \geq 1$
figures per image)\end{tabular}}} & \multicolumn{2}{l}{\begin{tabular}[c]{@{}l@{}}Revised dataset\\ (1 figure per image)\end{tabular}} \\
 & \multicolumn{1}{l|}{} & Train & Test \\ \cline{2-4} 
 & \multirow{2}{*}{30714} & \multirow{2}{*}{17877} & \multirow{2}{*}{2417} \\
 &  &  &  \\ \hline
\end{tabular}%
}
\\
\caption{The PatFig dataset statistics}
\end{table}
\begin{table}[]
\resizebox{\columnwidth}{!}{%
\begin{tabular}{l|llllllll}
\hline
\multirow{2}{*}{} & \multicolumn{3}{c|}{Input} & \multicolumn{5}{c}{Metric} \\
 & Image & Title & \multicolumn{1}{l|}{Terms} & BLEU2 & BLEU4 & ROUGE & METEOR & CIDEr \\ \hline
\multirow{8}{*}{Short captions} & \multicolumn{8}{c}{with reference numerals} \\ \cline{2-9} 
 & + & - & \multicolumn{1}{l|}{-} & 0.5322 & 0.3851 & 0.3677 & 0.3336 & 0.3669 \\
 & + & + & \multicolumn{1}{l|}{-} & 0.5537 & 0.4206 & 0.4007 & 0.3753 & 0.7947 \\
 & + & + & \multicolumn{1}{l|}{+} & 0.2202 & 0.1553 & 0.2358 & 0.2232 & 0.1903 \\ \cline{2-9} 
 & \multicolumn{8}{c}{without reference numerals} \\ \cline{2-9} 
 & + & - & \multicolumn{1}{l|}{-} & 0.5359 & 0.3922 & 0.4071 & 0.3640 & 0.3413 \\
 & + & + & \multicolumn{1}{l|}{-} & 0.5573 & 0.4276 & 0.4390 & 0.4056 & 0.7939 \\
 & + & + & \multicolumn{1}{l|}{+} & 0.2228 & 0.1577 & 0.2577 & 0.2338 & 0.1105 \\ \hline
\multicolumn{1}{c|}{\multirow{8}{*}{Long Captions}} & \multicolumn{8}{c}{with reference numerals} \\ \cline{2-9} 
\multicolumn{1}{c|}{} & + & - & \multicolumn{1}{l|}{-} & 0.3281 & 0.1936 & 0.1949 & 0.1410 & 0.0114 \\
\multicolumn{1}{c|}{} & + & + & \multicolumn{1}{l|}{-} & 0.3437 & 0.2162 & 0.2212 & 0.1595 & 0.0366 \\
\multicolumn{1}{c|}{} & + & + & \multicolumn{1}{l|}{+} & 0.3255 & 0.2166 & 0.2308 & 0.1740 & 0.0582 \\ \cline{2-9} 
\multicolumn{1}{c|}{} & \multicolumn{8}{c}{without reference numerals} \\ \cline{2-9} 
\multicolumn{1}{c|}{} & + & - & \multicolumn{1}{l|}{-} & 0.3313 & 0.1974 & 0.2317 & 0.1534 & 0.0142 \\
\multicolumn{1}{c|}{} & + & + & \multicolumn{1}{l|}{-} & 0.3478 & 0.2210 & 0.2598 & 0.1726 & 0.0418 \\
\multicolumn{1}{c|}{} & + & + & \multicolumn{1}{l|}{+} & 0.3267 & 0.2177 & 0.2552 & 0.1755 & 0.0587 \\ \hline 
\end{tabular}%
}
\label{tab:my_table3}
\\
\caption{Results for various captioning configurations, assessed both with and without reference numerals. Experiments using terms were done on test subsets with retrievable non-empty term lists.}
\end{table}
\subsection{Vision only}
\textbf{Task}: Given an image and general prompt generate the description

\textbf{Prompt 1}: \textit{Please provide the detailed description of the figure}.

The experiment consists of two main components: the input image and a simple prompt. The input image is fed into the model for processing, and the model generates a caption of the image based solely on its visual content. The goal of this experiment is to evaluate the model's ability to understand and describe images without any additional text-based context.

\subsection{Vision+Text}
\textbf{Task}: Given an image and a prompt including the patent title generate the description.

\textbf{Title}: \enquote{Activation of energy devices} 

\textbf{Terms}: 137602: \enquote{sensor}, 137604: \enquote{wired connection}, 137650: \enquote{surgical site opening}, 120: \enquote{patient side cart}, 137606: \enquote{surgical instrument}, 137600: \enquote{retractor}.

\textbf{Prompt 2}: \textit{Please provide the detailed description of the figure associated with \textbf{title}}.

\textbf{Prompt 3}: \textit{Please provide the detailed description of the figure associated with \textbf{title}. The image contains the following reference numerals and  \textbf{terms}}.

This experiment aims to evaluate the impact of added text-based cues on generating contextually accurate captions. The model uses the image as visual input and incorporates the title and terms as additional text-based context for more detailed and relevant captioning. The terms are retrieved from the patent application's complete description, and a subset of 500 samples from the test data is selected to assess the terms' effect.


\section{Discussion}

Long descriptions usually mention the different parts that the numbers in the figure refer to; short captions generally do not. As expected, the long caption generation generally benefits from adding the terms to the input, in particular for the CIDEr score, conceived as a caption metric (except for the BLEU2 score). Interestingly, the short caption generation seems to be disturbed by the term list in the input. This is in line with results by \cite{hsu2021scicap} 
on scientific image captioning. Their BLEU scores are similarly low, e.g. at 0.0231 for vision-only models, and decrease even further when adding textual cues. This might be explained by the fact that the model may overlook crucial visual features, resulting in less accurate captions. Additionally, conflicts between textual and visual cues may also confuse the model. 

It is well-known that neural vision models do not implicitly learn OCR, and with the goal of generating good captions, it is not actually necessary that our model generates the reference numerals as we could simply add the extracted reference numerals to the matching terms in the generated caption. So the model's performance is generally better when evaluating without taking the numerals into account.

The significant difference between the gold corpus and the generated captions can be attributed to limitations of MiniGPT-4: 1) The use of a frozen Q-former in the visual encoder may result in the loss of key features like visual-spatial grounding. 2) Training only a single projection layer may limit the model's ability to learn comprehensive visual-text alignment effectively.

   

\section{Conclusion and future work}
This paper introduces the first extensive dataset for gauging the efficiency of Large Vision Language Models (LVLM) on patent figures. Distinguished by reference numerals, a formal template-like caption style, and a wealth of text data linked to each figure, this dataset provides a unique challenge, differentiating it from conventional captioning tasks. Moreover, PatFig encapsulates wider types of patent images compared to the existing datasets, spanning technical drawings, block diagrams, flow charts, plots, and grayscale photographs. It offers multiple data points that can be harnessed for image captioning and potentially for other tasks such as image search and image generation, as well as addressing a broader scope of patent figure analysis tasks.

We delved into a key function of LVLM, exploring the dynamic interplay between language and image during the generation of two distinct caption types: short and long. These variants necessitate different input information, and our findings affirm that our LVLM models can effectively assimilate this information, thereby enhancing the results. Yet, various variations of our experiments should be studied, such as variations of training data size, prompt improvements, with the inclusion of textual cues during the finetuning stage, training directly on texts from which reference numerals have been removed, and, more difficult, removing the numbers from the images. Additionally, the identified patent figure types can be used to categorize the results based on each figure type. 

An intriguing aspect warranting further study is identifying the threshold where the image itself becomes redundant in the text generation process. In other words, discerning when a text-only large language model can accurately predict the figure's content without directly analyzing it. 

Future research will expand to investigate the generation of figures from patent text. This could not only streamline the work of patent attorneys significantly but also shed light on the necessary information for drawing a figure and how this data is amalgamated to create a figure.
\section{Aknowledgments}
The authors would like to thank the rest of the Qatent team, including all researchers, engineers, developers, and law experts, for their insights and collaboration throughout the project. 

{\small
\bibliographystyle{ieee_fullname}
\bibliography{egbib}
}
\pagebreak

\appendix
\section{Figure type extraction}
\label{appendix:fig}
\begin{figure}[!h]
\begin{center}
   \includegraphics[width=0.8\linewidth]{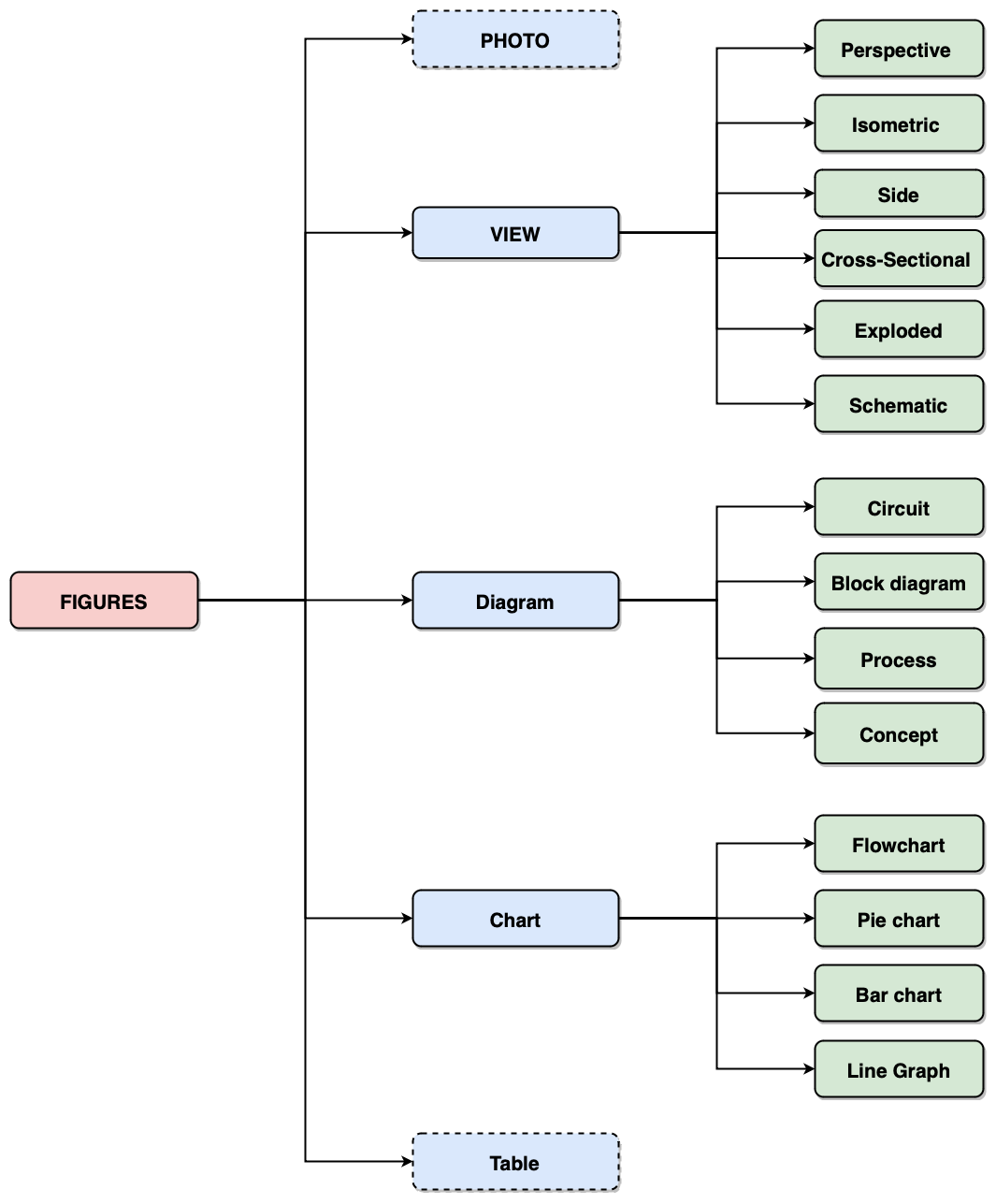}
\end{center}
   \caption{The taxonomy of patent figure types.}
\label{fig:long}
\label{fig:onecol}
\end{figure}

\section{Model predictions}\label{appendix:pred}
\begin{table*}[!ht]
    \centering
    \small
    
    \begin{tabularx}{\linewidth}{|X|X|}
        \hline
        Patent number & EP3554181B1 \\
        \hline
        Image id & EP\_3554181\_B1\_4.png \\
        \hline
        Title & FRONT-HAUL TRANSPORT NETWORK, DATA TRANSMISSION METHOD, APPARATUS AND COMPUTER STORAGE MEDIUM \\
        \hline
        Terms & 1201: \enquote{selecting unit}, 1202: \enquote{first sending unit} \\
        \hline
        Minimal claim set 
        
        (Claims covering the maximum number of relevant terms) & 
            15. A device for data transmission, characterized in that it is configured to control data transmission of a front-haul transport network, FTN, wherein the FTN comprises: an access-type front-haul transport node, FTN-ACC, connected to a remote radio unit, RRU; and an aggregation-type front-haul transport node, FTN-AGG, connected to the FTN-ACC and a base band unit, BBU, pool respectively, wherein the FTN-ACC is a node separate from the RRU and the FTN-AGG is a node separate from the BBU pool; wherein the device for data transmission is the FTN-ACC or the FTN-AGG, and comprises: a selecting unit (1201), configured to select, according to a delay requirement of data, a transmission channel established between the FTN-ACC and the FTN-AGG; and a first sending unit (1202), configured to send the data by using the transmission channel.
        \\
        \hline
        Short Caption & Long Caption \\
        \hline
        \multicolumn{2}{|c|}{Groundtruth}  \\
        \hline
        FIG12 illustrates a schematic structural diagram of a device for data transmission of an FTN according to an example of the present disclosure & A device for data transmission is further provided in this example, which is configured to control data transmission of an FTN. The FTN includes an FTN-ACC connected to the RRU, and an FTN-AGG connected to the FTN-ACC and a BBU pool respectively. As illustrated in FIG12, the device for data transmission includes a selecting unit 1201 and a first sending unit 1202. \\
        \hline
        \multicolumn{2}{|c|}{Predicted: Image}  \\
        \hline
         FIG12 is a schematic structural diagram of a terminal according to an embodiment of the present invention & In an embodiment of the present invention, the network device may be a communication device. The communication device may be a base station, a relay station, or a user equipment. FIG12 is a schematic structural diagram of a communication device according to an embodiment of the present invention. As shown in FIG12, the communication device may include a processor 1201, a memory 1202, and a transceiver 1203. The processor 1201 may be configured to implement the foregoing methods of the embodiments of the present invention. The memory 1202 may be configured to store the foregoing methods of the embodiments of the present invention. The transceiver 1203 may be configured to send and receive signals. \\
        \hline
        \multicolumn{2}{|c|}{Predicted: Image + Title}  \\
        \hline
        FIG13 is a schematic structural diagram of a base station according to an embodiment of the present invention & FIG1 is a schematic structural diagram of a front-haul transport network according to an embodiment of the present invention. As shown in FIG1, the front-haul transport network may include a first base station and a second base station. The first base station and the second base station may be connected to each other through a first optical fiber and a second optical fiber, respectively. The first optical fiber and the second optical fiber may be connected to each other through a third optical fiber. The first base station and the second base station may be connected to each other through the third optical fiber. \\
        \hline
        \multicolumn{2}{|c|}{Predicted: Image + Title + Terms}  \\
        \hline
        This is a diagram of a front-haul transport network and a data transmission method, apparatus, and computer storage medium thereof according to an embodiment of the present invention & FIG12 is a schematic structural diagram of an apparatus for transmitting data according to an embodiment of the present invention. As shown in FIG12, the apparatus 1200 for transmitting data may include a selecting unit 1201 and a first sending unit 1202. \\
        \hline 
        \multicolumn{2}{c}{} \\
    \end{tabularx}
    
    \caption{Test set data sample with associated title, patent number, a minimal claim set, ground truth, and predicted short and long captions.}
    \label{tab:my_table4}

\end{table*}

\end{document}